\documentclass[letterpaper, 10 pt, conference]{ieeeconf}  

\IEEEoverridecommandlockouts                              

\usepackage{cite}
\usepackage{amsmath,amssymb,amsfonts}
\usepackage{algorithmic}
\usepackage{graphicx}
\usepackage{textcomp}
\usepackage{xcolor}
\usepackage{hyperref}
\usepackage{authblk}
\usepackage{amssymb}
\usepackage{booktabs}
\usepackage{multirow}
\usepackage{pifont}
\usepackage{subcaption}
\usepackage{adjustbox} 
\usepackage{multirow}  
\usepackage{amsfonts}
\hypersetup{
    colorlinks=true,
    linkcolor=blue, 
    citecolor=blue, 
    urlcolor=purple 
}
\usepackage{orcidlink}
\usepackage[capitalize]{cleveref}
\crefname{section}{Sec.}{Secs.}
\Crefname{section}{Section}{Sections}
\Crefname{table}{Table}{Tables}
\crefname{table}{Table}{Tables}
\Crefname{figure}{Figure}{Figures}
\crefname{figure}{Fig.}{Figs.}
\Crefname{equation}{Equation}{Equations}
\crefname{equation}{Eq.}{Eqs.}
\crefname{algocf}{alg.}{algs.}
\Crefname{algocf}{Algorithm}{Algorithms}
\usepackage{xspace}
\makeatletter
\DeclareRobustCommand\onedot{\futurelet\@let@token\@onedot}
\def\@onedot{\ifx\@let@token.\else.\null\fi\xspace}

\def\etal{\emph{et al}\onedot}
\makeatother
\def\BibTeX{{\rm B\kern-.05em{\sc i\kern-.025em b}\kern-.08em
    T\kern-.1667em\lower.7ex\hbox{E}\kern-.125emX}}
\begin{document}

\title{MAC-Lookup: Multi-Axis Conditional Lookup Model for Underwater Image Enhancement
\thanks{* Equal contribution.}
\thanks{$^\dag$ Corresponding author.}
}

\author[1*]{Fanghai Yi}
\author[1*]{Zehong Zheng}
\author[2]{Zexiao Liang}
\author[3]{Yihang Dong}
\author[2]{Xiyang Fang}
\author[4]{Wangyu Wu}
\author[2$^\dag$\orcidlink{0000-0001-6000-3914}]{Xuhang Chen}

\affil[1]{Guangdong University of Technology}
\affil[2]{School of Computer Science and Engineering, Huizhou University}
\affil[3]{University of Chinese Academy of Sciences}
\affil[4]{Xi'an Jiaotong-Liverpool University}
\maketitle

\begin{abstract}
Enhancing underwater images is crucial for exploration. These images face visibility and color issues due to light changes, water turbidity, and bubbles. Traditional prior-based methods and pixel-based methods often fail, while deep learning lacks sufficient high-quality datasets. We introduce the Multi-Axis Conditional Lookup (MAC-Lookup) model, which enhances visual quality by improving color accuracy, sharpness, and contrast. It includes Conditional 3D Lookup Table Color Correction (CLTCC) for preliminary color and quality correction and Multi-Axis Adaptive Enhancement (MAAE) for detail refinement. This model prevents over-enhancement and saturation while handling underwater challenges. Extensive experiments show that MAC-Lookup excels in enhancing underwater images by restoring details and colors better than existing methods. The code is \url{https://github.com/onlycatdoraemon/MAC-Lookup}.
\end{abstract}

\section{Introduction}
Advancements in underwater technology have made image enhancement crucial for exploring oceanic environments. However, factors like light variations, water turbidity, and bubbles degrade image quality, leading to visibility and color issues that complicate image analysis~\cite{li1,li2,li3,wu2025image}.

Underwater image enhancement methods split into traditional and learning-based categories. Initially, they used priors to clarify images, but over-reliance weakened their effectiveness and robustness~\cite{akkaynak2019sea}. In contrast, enhancement-based methods adjust pixel values to boost contrast and brightness without priors but risk over-enhancement or over-saturation~\cite{song2021enhancement}. Recently, deep learning~\cite{li2019underwater,wu2025generative,peng2023u,liu2023coordfill,zhang2022correction,li2025adaptive,li2023cee,li2022monocular,li2022few} gained prominence due to extensive datasets, yet lacks of high-quality, large-scale paired images hinders them. The dynamic underwater environment further complicates performance. Thus, integrating various method strengths is vital to enhance underwater image quality and technique efficiency.
\begin{figure}
    \centering
    \includegraphics[width=1\linewidth]{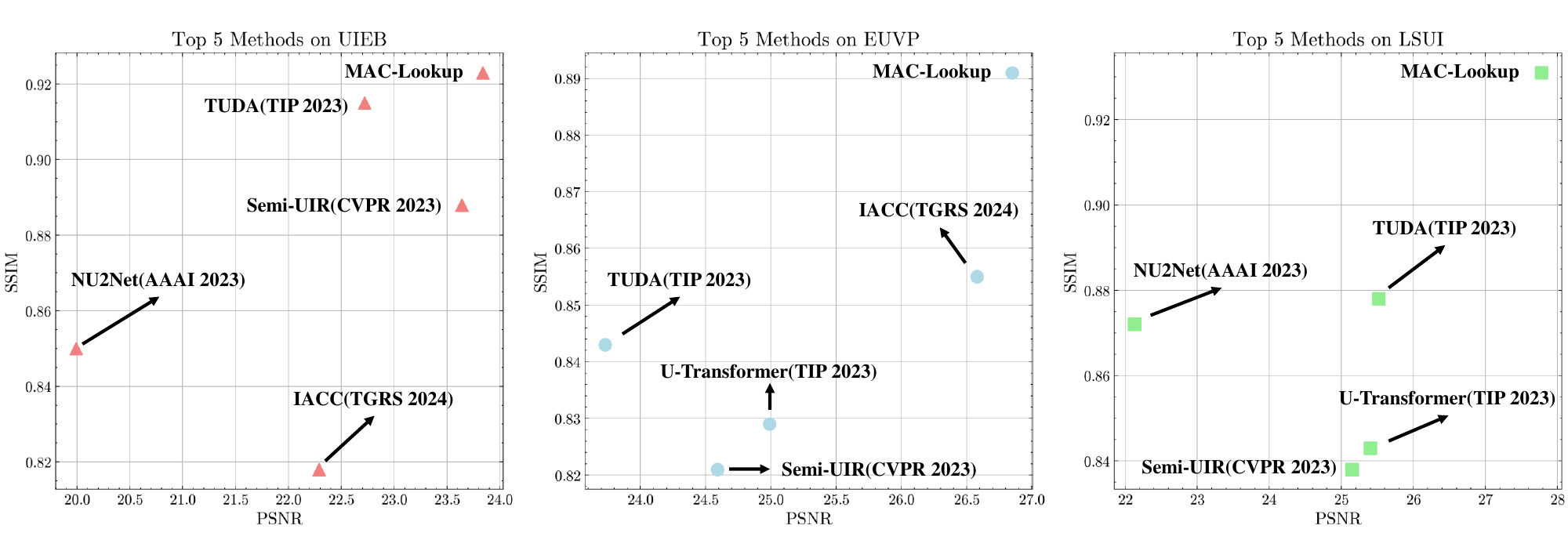}
    \caption{Top-5 performance comparison on UIEB~\cite{li2019underwater}, EUVP~\cite{islam2020fast}, and LSUI~\cite{peng2023u} datasets.}
    \label{fig:eperformance}
\end{figure}

We introduce the MAC-Lookup model to improve underwater image quality by enhancing color accuracy, sharpness, and contrast. It consists of the Conditional 3D Lookup Table Color Correction (CLTCC) for initial correction and the Multi-Axis Adaptive Enhancement (MAAE) for further refinement. The model learns by comparing ground-truth with CLTCC-processed images, mitigating over-enhancement and saturation while balancing global and local detail.
\begin{itemize} 
\item We propose an efficient method called MAC-Lookup, which demonstrates superior performance in the underwater image enhancement task.
\item We introduce a Conditional 3D Lookup Table for color correction that minimizes color loss by accounting for the attenuation characteristics of different color channels, offering an improvement over traditional color correction techniques. 
\item We propose a Multi-Axis adaptive enhancement technique that uses local and global branches to accurately predict images, thereby reducing unnatural region and the issues of under and over-enhancement prevalent in existing methods. 
\end{itemize}

\section{Related Work}

\subsection{Traditional Underwater Image Enhancement}
Traditional methods for enhancing underwater images often use physical models and image processing to mitigate light absorption and scattering. Akkaynak \etal~\cite{akkaynak2019sea} developed Sea-thru, which uses a precise image formation model to remove water effects by estimating range-dependent parameters. Zhuang \etal~\cite{zhuang2022underwater} introduced a hyper-Laplacian reflectance prior retinex model for better visibility and less color distortion by accurately estimating reflectance and illumination. Zhang \etal~\cite{zhang2022underwater} proposed MLLE, which improves color, contrast, and details with minimal color loss and adaptive contrast enhancement. Liu \etal~\cite{liu2022rank} created a real-time recovery framework, ROP and ROP+, to restore degraded images. Zhang \etal~\cite{zhang2023underwater} developed WWPF, using wavelet perception fusion to correct color distortion and enhance contrast, validated through assessments. Hou \etal~\cite{hou2023non} introduced an ICSP-guided variational framework with a fast algorithm and NUID benchmark dataset for restoring non-uniform illumination.

\subsection{Learning-Based Underwater Image Enhancement}
Learning-based methods~\cite{liu2023explicit,zhu2024test,liu2024depth,liu2024dh,li2024cross,liu2024forgeryttt,zheng2024smaformer,jiang2021deep}, using deep neural networks, are becoming essential in underwater image enhancement. Li \etal~\cite{li2019underwater} introduced the UIEB dataset with 950 real underwater images and Water-Net, a CNN for enhancing these images. Cong \etal~\cite{cong2023pugan} developed PUGAN, a GAN that uses a dual-discriminator framework focusing on color correction and visual quality. Wang \etal~\cite{wang2023domain} presented TUDA, which uses a dual-alignment network for minimizing domain gaps, and rank-based image quality assessment. Guo \etal~\cite{guo2023underwater} created URanker, a ranking-based assessment tool employing a conv-attentional Transformer and a new dataset, URankerSet, to advance UIE networks. Peng \etal~\cite{peng2023u} developed the U-shape Transformer for addressing uneven color attenuation and introduced a new large-scale dataset for UIE. Huang \etal~\cite{huang2023contrastive} presented Semi-UIR, a semi-supervised framework using the mean-teacher approach with contrastive regularization to reduce overfitting. Zhou \etal~\cite{zhou2024iacc} proposed IACC, enhancing underwater images under mixed lighting by aligning luminance features and correcting color using complementary colors. There are other methods based on transformer and LVLMs \cite{zhang1,zhang2,zhang3,zhang4,zhang5,zhang6,zhang7,zhang8,zhang9,zhang10,zhang11,zhang12,zhang13,zhang14} that inspire us.

\begin{figure}[ht]
    \centering
    \includegraphics[width=1\linewidth]{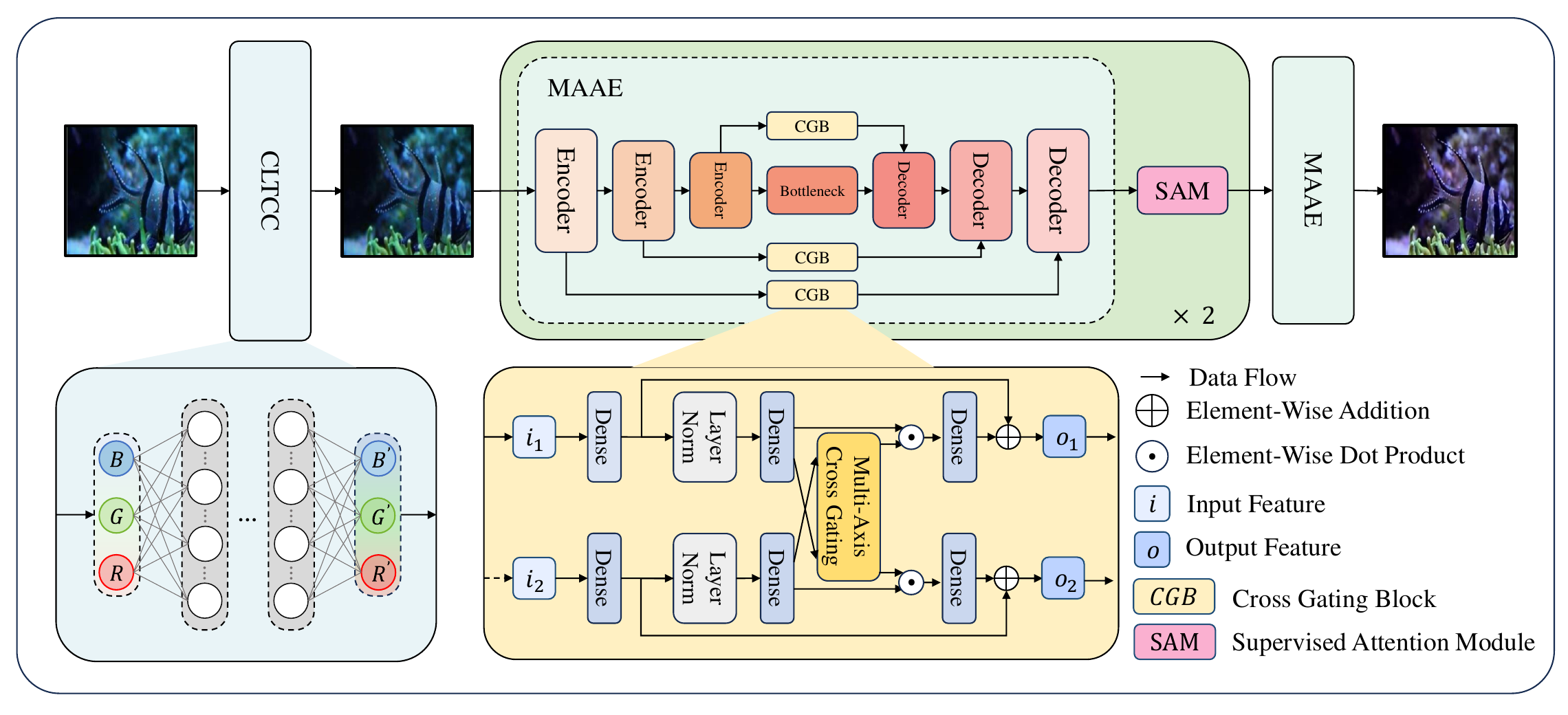}
    \caption{Overview of the proposed MAC-Lookup framework.}
    \label{fig:whole}
\end{figure}

\section{Methodology}
We present the Multi-Axis Conditional Lookup Model (MAC-Lookup) for underwater image processing, as shown in \cref{fig:whole}. Unlike previous methods~\cite{akkaynak2019sea,zhuang2022underwater,liu2022rank} relying on physical models, MAC-Lookup uses deep learning to tackle complex underwater environments and over-enhancement issues. It comprises two components: Conditional 3D Lookup Table Color Correction (CLTCC) and Multi-Axis Adaptive Enhancement (MAAE). CLTCC provides initial light and color corrections with linear complexity. MAAE utilizes a multi-axis design to balance local (Conv) and global (MLP) modules, enhancing detail restoration and natural color transitions. The Cross Gating Block (CGB) improves encoder-decoder interaction for better global consistency. Post-CLTCC, multi-stage MAAE further refines image quality using Supervised Attention Blocks (SAM) for feature extraction across stages. This design surpasses current best results without extensive pre-training. The complete model structure is illustrated in \cref{fig:whole}.

\subsection{Conditional 3D Lookup Table Color Correction}
Due to light absorption in water, underwater images often exhibit significant blue or green color casts. Therefore, color correction is essential to improve visibility and restore the original appearance of underwater scenes. Inspired by NILUT~\cite{conde2024nilut}, which employs 3D lookup tables (LUTs) in combination with a neural implicit network to modify image color styles, we aim to learn a nonlinear 3D color transformation for underwater image enhancement. Specifically, NILUT approximates this transformation by sparsely sampling a 3D lattice. Following this concept, our objective is to develop a 3D transformation model that maps underwater images to their ground-truth counterparts in the RGB color space. This is achieved by learning a continuous function, denoted as $\Psi$, which maps $\mathbb{R}^3$ coordinates from an input image $\textbf{I}$ to a target image $\textbf{T}$, both of which reside in 3D domains:
\begin{equation}
\Psi: \mathbb{R}^3 \to \mathbb{R}^3 \quad \text{with} \quad \Psi(\textbf{I}) \in [0, 1]^3.
\end{equation}

The output of $\Psi$ is represented as $\Phi$:
\begin{equation}
\Phi = \Psi(\textbf{I}).
\end{equation}

The function $\Psi$ serves as an implicit neural representation of a 3D lookup table and is formulated as:
\begin{equation}
\begin{split}
    \Psi(x) &= \mathcal{W}_n(\xi_{n-1} \circ \xi_{n-2} \circ \dots \circ \xi_{0})(x) + b_n, \\
    \xi_i &= \alpha(\mathcal{W}_i x_i + b_i),
\end{split}
\end{equation}
where $\xi_i$ represents the layers of the network (with their corresponding weight matrix $\mathcal{W}_i$ and bias $b_i$), and $\alpha$ is a nonlinear activation function.

\subsection{Multi-Axis Adaptive Enhancement}
Although the CLTCC module performs preliminary corrections to the colors and details of underwater images, issues such as partial color deviations, regional blurriness, and insufficient illumination remain. Additionally, we observed that existing methods for underwater image enhancement, such as FUnIE-GAN~\cite{islam2020fast}, suffer not only from over-saturation and over-enhancement but also from unnatural transitions in certain regions, characterized by color discontinuities and abrupt changes.

We hypothesize that addressing these challenges requires a seamless integration of global and local perspectives. This ensures the meticulous restoration of local details while maintaining smooth and natural transitions between regions. Inspired by MAXIM~\cite{tu2022maxim}, we adopt a multi-axis approach, leveraging multi-axis gated MLPs to efficiently and scalably blend local and global visual cues. Furthermore, we incorporate cross-gated blocks, an alternative to cross-attention, to enhance global contextual features that gate skip connections.

This approach aligns perfectly with our objectives, achieving superior performance with fewer parameters and faster computational speeds. Based on this foundation, we developed a novel component, MAAE, within the MAXIM framework, with its detailed structure illustrated in \cref{fig:whole}.

To further improve image restoration, we stack multiple MAAE units, using a Supervised Attention Mechanism (SAM) between them to enhance feature extraction at each stage.

Formally, given an input image $\mathbf{I} \in \mathbb{R}^{H \times W \times 3}$, the processed images after CLTCC corrections are fed into MAAE and SAM as follows:
\begin{equation}
  \begin{split}
    O_1 &= \text{MAAE}_1(\mathbf{I}), \\
    O_i &= \text{MAAE}_i(\text{SAM}(O_{i-1})), \quad i \geq 2,
  \end{split}  
\end{equation}
where $O_i$ represents the restored output at each stage, and $\text{MAAE}_i$ denotes the MAAE unit at stage $i$. At each stage $\mathcal{S}$, MAAE predicts multi-scale restored outputs $\mathbf{I}_n$ for $n = 1, \dots, \mathcal{N}$, producing $\mathcal{S} \times \mathcal{N}$ outputs denoted as $O_{s,n}$.

\subsection{Objective Function}
For CLTCC optimization, the RGB space is represented as a set of coordinates $\textit{X} = \{(r_i, g_i, b_i)\}$. To learn the continuous representations $\Phi$, we minimize the following loss function:
\begin{equation}
\mathcal{L}_{CLTCC} = \sum_i \|\Phi(x_i) - \textbf{T}(x_i)\|_1,
\end{equation}
where $\textbf{T}$ denotes the real conditional vector. As for the loss function for MAAE is defined as:
\begin{equation}
    \mathcal{L}_{MAAE} = \sum_{s=1}^{S} \sum_{n=1}^{N} \left[ \mathcal{L}_{\text{char}}(\mathbf{O}_{s,n}, \mathbf{T}_n) + \lambda \mathcal{L}_{\text{freq}}(\mathbf{O}_{s,n}, \mathbf{T}_n) \right],
\end{equation}
where $\mathbf{T}_n$ denotes bilinearly-rescaled multi-scale target images, and $\mathcal{L}_{\text{char}}$ is the Charbonnier loss:
\begin{equation}
    \mathcal{L}_{\text{char}}(\mathbf{O}, \mathbf{T}) = \sqrt{\|\mathbf{O} - \mathbf{T}\|^2 + \epsilon^2},
\end{equation}
with $\epsilon = 10^{-3}$ as in MAXIM. The frequency reconstruction loss, $\mathcal{L}_{\text{freq}}$, enforces high-frequency detail preservation:
\begin{equation}
    \mathcal{L}_{\text{freq}}(\mathbf{O}, \mathbf{T}) = \|\mathcal{F}(\mathbf{O}) - \mathcal{F}(\mathbf{T})\|_1,
\end{equation}
where $\mathcal{F}(\cdot)$ represents the 2D Fast Fourier Transform. Following MAXIM, we set the weighting factor $\lambda = 0.1$ in all experiments.

Next, we evaluate the quality of the generated images. We selected PSNR and SSIM as the metrics to assess the quality of image restoration. The evaluation is performed using the output of the last stage and the final resolution of MAC-Lookup as the model output. The evaluation loss $\mathcal{L}_{GT}$ is defined as follows:
\begin{equation}
    \begin{split}
        &\mathcal{L}_{GT} = \mathcal{L}_{PSNR} + 0.4\times\mathcal{L}_{SSIM}, \\
        &\mathcal{L}_{PSNR} = PSNR(O, T), \\
        &\mathcal{L}_{SSIM} = 1 - SSIM(O, T),
    \end{split}
\end{equation}
where $O$ denotes the output image, and $T$ denotes the ground truth image. Finally, we define the overall loss function for the training process as:
\begin{equation}
    \mathcal{L}_{total} = \mathcal{L}_{GT} + 0.5\times\mathcal{L}_{CLTCC} + 0.5\times\mathcal{L}_{MAAE},
\end{equation}
where $\mathcal{L}_{CLTCC}$ and $\mathcal{L}_{MAAE}$ represent the losses associated with the CLTCC and MAAE modules, respectively.

\begin{table*}[ht]
\centering
\caption{Comparisons on UIEB~\cite{li2019underwater}, EUVP~\cite{islam2020fast}, and LSUI~\cite{peng2023u} dataset. The best result is highlighted in bold while the second best result is highlighted in underline.}
\label{tab:experiment}
\adjustbox{max width=\linewidth}{
\begin{tabular}{l c c cc cc cc cc c c}
\toprule
\multirow{3}{*}{\textbf{Method}}  & 

\multicolumn{3}{c}{\textbf{UIEB (90 images)}} & \multicolumn{3}{c}{\textbf{EUVP (200 images)}} & \multicolumn{3}{c}{\textbf{LSUI (200 images)}}&
\multirow{3}{*}{\textbf{Params (M)}}&
\multirow{3}{*}{\textbf{GMACs}}\\
\cmidrule(lr){2-4} \cmidrule(lr){5-7} \cmidrule(lr){8-10}&   \textbf{PSNR}$\uparrow$& \textbf{SSIM}$\uparrow$& \textbf{LPIPS}$\downarrow$& \textbf{PSNR}$\uparrow$& \textbf{SSIM}$\uparrow$& \textbf{LPIPS}$\downarrow$& \textbf{PSNR}$\uparrow$& \textbf{SSIM}$\uparrow$& \textbf{LPIPS}$\downarrow$\\
\midrule

WCID \cite{chiang2011underwater} & 11.65 & 0.322 & 0.424 & 12.93 & 0.270 & 0.480 & 12.38 & 0.285 & 0.444 &- &-  \\ 
Sea-thru \cite{akkaynak2019sea} & 13.82 & 0.580 & 0.421 & 12.72 & 0.499 & 0.496 & 12.91 & 0.505 & 0.501 &- &-\\
CIEUI \cite{sethi2019adaptive} & 14.47 & 0.747 & 0.364 & 15.46 & 0.723 & 0.410 & 14.82 & 0.768 & 0.408 &- &-\\
FUnIE-GAN \cite{islam2020fast} & 19.17 & 0.800 & 0.217 & 22.62 & 0.720 & 0.230 & 21.66 & 0.744 & 0.240 & 7.7 & 10.7\\ 
Bayesian-retinex \cite{zhuang2021bayesian} & 18.75 & 0.829 & 0.262 & 15.64 & 0.669 & 0.369 & 17.59 & 0.747 & 0.321 &- &-\\
SFGNet \cite{zhao2024toward}& 19.57 & 0.685 & 0.214 & 22.68 & 0.585 & 0.221 & 22.71 & 0.653 & 0.204 & \textbf{1.2}& 81.5\\
IACC \cite{zhou2024iacc}& 22.29 & 0.818 & 0.180 & \underline{26.58} & \underline{0.855} & \underline{0.159} & 23.62 & 0.678 & 0.175 & \underline{2.1} & 132.4\\
TUDA \cite{wang2023domain} & 22.72 & \underline{0.915} & \underline{0.118} & 23.73 & 0.843 & 0.207 & \underline{25.52} & \underline{0.878} & \underline{0.154} &2.7 &85.4\\ 
PUGAN \cite{cong2023pugan}& 20.52 & 0.812 & 0.216 & 22.58 & 0.820 & 0.212 & 23.14 & 0.836 & 0.216 &- &-\\ 
U-Transformer \cite{peng2023u} & 20.75 & 0.810 & 0.228 & 24.99 & 0.829 & 0.238 & 25.15 & 0.838 & 0.221 & 2.9 & \underline{22.8}\\
NU2Net \cite{guo2023underwater} & 19.99 & 0.850 & 0.196 & 21.51 & 0.810 & 0.308 & 22.13 & 0.872 & 0.243  &3.1  &\textbf{10.4}\\
Semi-UIR \cite{huang2023contrastive} & \underline{23.64} & 0.888 & 0.120 & 24.59 & 0.821 & 0.172 & 25.40 & 0.843 & 0.160 &- &-\\
ICSP \cite{hou2023non} & 12.04 & 0.599 & 0.552 & 11.73 & 0.522 & 0.413 & 11.96 & 0.583 & 0.508 &- &-\\
HLRP \cite{zhuang2022underwater}& 13.30 & 0.259 & 0.364 & 11.41 & 0.186 & 0.500 & 12.96 & 0.221 & 0.429 &- &-\\
MLLE \cite{zhang2022underwater} & 18.74 & 0.814 & 0.234 & 15.14 & 0.633 & 0.323 & 17.87 & 0.730 & 0.278 &- &-\\
ROP \cite{liu2022rank} & 18.48 & 0.849 & 0.209 & 15.34 & 0.714 & 0.343 & 17.38 & 0.806 & 0.281 &- &-\\

\midrule
MAC-Lookup& \textbf{23.84} & \textbf{0.923} & \textbf{0.103} & \textbf{26.85} & \textbf{0.891} & \textbf{0.139} & \textbf{27.78} & \textbf{0.931} & \textbf{0.099} & 26.8 & 173.2\\
\bottomrule
\end{tabular}}
\end{table*}

\section{Experiment}
\subsection{Dataset}
Underwater image datasets are divided into synthetic and real-world types, with models trained on synthetic data often struggling with real scenes. To address this, we used high-quality real-world datasets, sampling 800 images from UIEB~\cite{li2019underwater}, 2000 from EUVP~\cite{islam2020fast}, and 2000 from LSUI~\cite{peng2023u} for training. For evaluation, we compiled test sets of 90 UIEB, 200 EUVP, and 200 LSUI images, ensuring no overlap with the training set.

\subsection{Evaluation Metrics}
\label{sec:metrics}
We employ multiple complementary metrics for a comprehensive performance assessment:
\begin{itemize}
    \item \textbf{PSNR} (Peak Signal-to-Noise Ratio): Measures the overall image quality by evaluating the reconstruction error between the original and reconstructed images. Higher PSNR values generally indicate better image quality.
    \item \textbf{SSIM}~\cite{wang2004image} (Structural Similarity Index): Evaluates the structural fidelity of images by assessing luminance, contrast, and structural information, thereby aligning more closely with human visual perception.
    \item \textbf{LPIPS}~\cite{zhang2018unreasonable} (Learned Perceptual Image Patch Similarity): Quantifies perceptual quality using deep learning models to capture high-level feature differences, providing a measure that better reflects human subjective judgment of image quality.
\end{itemize}
Superior performance is indicated by higher PSNR and SSIM values, as well as lower LPIPS scores.

\subsection{Implementation Details}
Our model is implemented in PyTorch and trained on dual NVIDIA A6000 GPUs. Training employs $256 \times 256$ input images over 200 epochs with a batch size of 4 using the Adam optimizer (initial learning rate $2 \times 10^{-4}$, minimum learning rate $1 \times 10^{-6}$). Data augmentation includes random cropping, flipping, rotation, and mix-up. Testing processes images at a resolution of $256 \times 256$.

\subsection{Comparisons with State-of-the-art Approaches}
We compared our network with top underwater image enhancement methods: WCID \cite{chiang2011underwater}, Sea-thru \cite{akkaynak2019sea}, CIEUI \cite{sethi2019adaptive}, FUnIE-GAN \cite{islam2020fast}, Bayesian-retinex \cite{zhuang2021bayesian}, SFGNet \cite{zhao2024toward}, IACC \cite{zhou2024iacc}, TUDA \cite{wang2023domain}, PUGAN \cite{cong2023pugan}, U-Transformer \cite{peng2023u}, NU2Net \cite{guo2023underwater}, Semi-UIR \cite{huang2023contrastive}, ICSP \cite{hou2023non}, HLRP \cite{zhuang2022underwater}, MLLE \cite{zhang2022underwater}, and ROP \cite{liu2022rank}.

As shown in \cref{tab:experiment}, MAC-Lookup outperforms other methods on the UIEB dataset. It matches Semi-UIR in PSNR and TUDA in SSIM and LPIPS, proving effective in solving color issues and preventing over-enhancement. MAC-Lookup's MAAE component enhances interactions in global and local branches, improving predictions. 

We evaluate MAC-Lookup's performance on the EUVP dataset, showing it notably surpasses state-of-the-art methods. As demonstrated in \cref{tab:experiment}, MAC-Lookup gains 0.27\% in PSNR, 0.36\% in SSIM, and -0.02\% in LPIPS for 200 tested images compared to IACC. Its effectiveness is due to noise mitigation and handling complex underwater conditions, allowing more focus on image details.

We assess MAC-Lookup on the LSUI dataset, as shown in \cref{tab:experiment}, where it outperforms all underwater image enhancement methods, including TUDA, with 2.26\% higher PSNR, 0.053\% higher SSIM, and 0.055\% lower LPIPS. This is due to its advanced architecture and MAAE, efficiently aggregating patterns across scales.

We also evaluate the balance between model complexity and performance. MAC-Lookup, despite its computational heaviness (\textbf{173.2M} parameters and \textbf{26.8 GMACs}), significantly outperforms lighter models like U-Transformer and NU2Net in all metrics, indicating its suitability for applications prioritizing visual quality over resources.

\begin{figure*}[ht]
    \begin{minipage}[b]{1.0\linewidth}
        \begin{minipage}[b]{0.075\linewidth}
            \centering
            \centerline{\includegraphics[width=\linewidth]{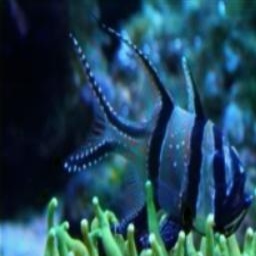}}
        \end{minipage}
        \hfill
        \begin{minipage}[b]{0.075\linewidth}
            \centering
            \centerline{\includegraphics[width=\linewidth]{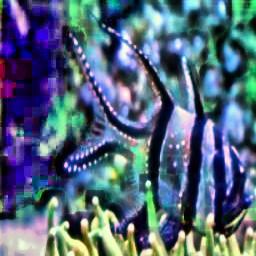}}
        \end{minipage}
        \hfill
        \begin{minipage}[b]{0.075\linewidth}
            \centering
            \centerline{\includegraphics[width=\linewidth]{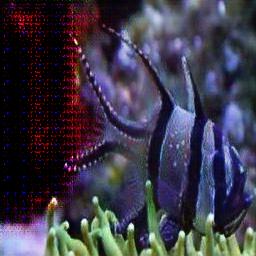}}
        \end{minipage}
        \hfill
        \begin{minipage}[b]{0.075\linewidth}
            \centering
            \centerline{\includegraphics[width=\linewidth]{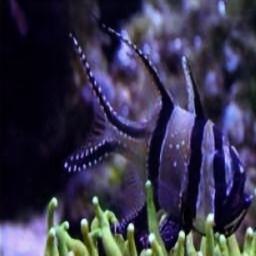}}
        \end{minipage}
        \hfill
        \begin{minipage}[b]{0.075\linewidth}
            \centering
            \centerline{\includegraphics[width=\linewidth]{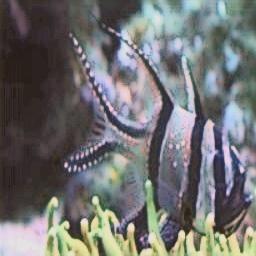}}
        \end{minipage}
        \hfill
        \begin{minipage}[b]{0.075\linewidth}
            \centering
            \centerline{\includegraphics[width=\linewidth]{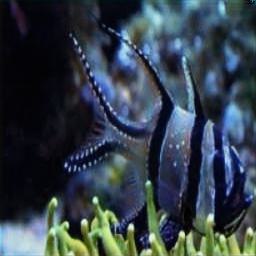}}
        \end{minipage}
        \hfill
        \begin{minipage}[b]{0.075\linewidth}
            \centering
            \centerline{\includegraphics[width=\linewidth]{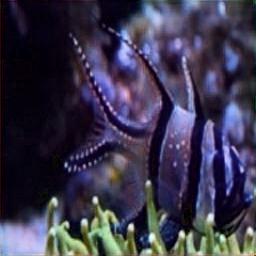}}
        \end{minipage}
        \hfill
        \begin{minipage}[b]{0.075\linewidth}
            \centering
            \centerline{\includegraphics[width=\linewidth]{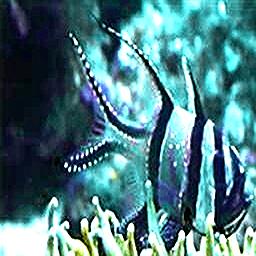}}
        \end{minipage}
        \hfill
        \begin{minipage}[b]{0.075\linewidth}
            \centering
            \centerline{\includegraphics[width=\linewidth]{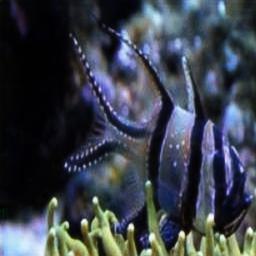}}
        \end{minipage}
        \hfill
        \begin{minipage}[b]{0.075\linewidth}
            \centering
            \centerline{\includegraphics[width=\linewidth]{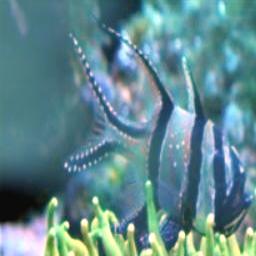}}
        \end{minipage}
        \hfill
        \begin{minipage}[b]{0.075\linewidth}
            \centering
            \centerline{\includegraphics[width=\linewidth]{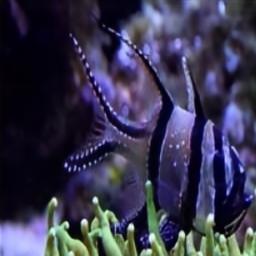}}
        \end{minipage}
        \hfill
        \begin{minipage}[b]{0.075\linewidth}
            \centering
            \centerline{\includegraphics[width=\linewidth]{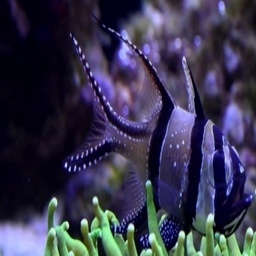}}
        \end{minipage}
    \end{minipage}

    \begin{minipage}[b]{1.0\linewidth}
        \begin{minipage}[b]{0.075\linewidth}
            \centering
            \centerline{\includegraphics[width=\linewidth]{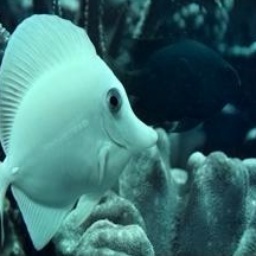}}
        \end{minipage}
        \hfill
        \begin{minipage}[b]{0.075\linewidth}
            \centering
            \centerline{\includegraphics[width=\linewidth]{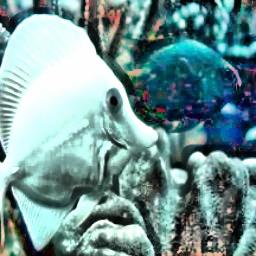}}
        \end{minipage}
        \hfill
        \begin{minipage}[b]{0.075\linewidth}
            \centering
            \centerline{\includegraphics[width=\linewidth]{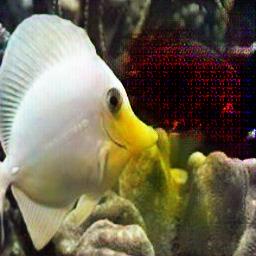}}
        \end{minipage}
        \hfill
        \begin{minipage}[b]{0.075\linewidth}
            \centering
            \centerline{\includegraphics[width=\linewidth]{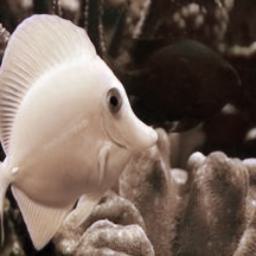}}
        \end{minipage}
        \hfill
        \begin{minipage}[b]{0.075\linewidth}
            \centering
            \centerline{\includegraphics[width=\linewidth]{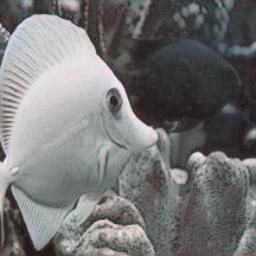}}
        \end{minipage}
        \hfill
        \begin{minipage}[b]{0.075\linewidth}
            \centering
            \centerline{\includegraphics[width=\linewidth]{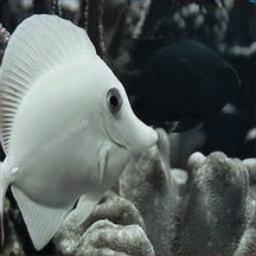}}
        \end{minipage}
        \hfill
        \begin{minipage}[b]{0.075\linewidth}
            \centering
            \centerline{\includegraphics[width=\linewidth]{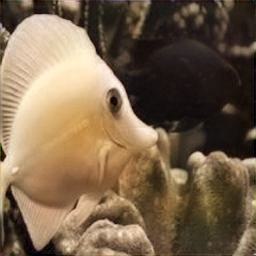}}
        \end{minipage}
        \hfill
        \begin{minipage}[b]{0.075\linewidth}
            \centering
            \centerline{\includegraphics[width=\linewidth]{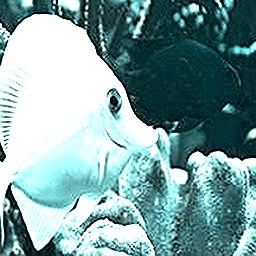}}
        \end{minipage}
        \hfill
        \begin{minipage}[b]{0.075\linewidth}
            \centering
            \centerline{\includegraphics[width=\linewidth]{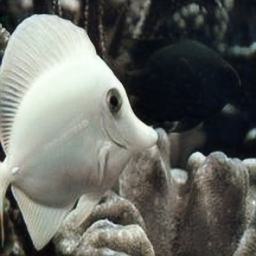}}
        \end{minipage}
        \hfill
        \begin{minipage}[b]{0.075\linewidth}
            \centering
            \centerline{\includegraphics[width=\linewidth]{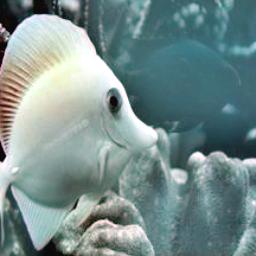}}
        \end{minipage}
        \hfill
        \begin{minipage}[b]{0.075\linewidth}
            \centering
            \centerline{\includegraphics[width=\linewidth]{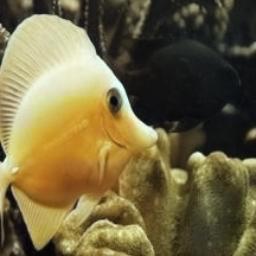}}
        \end{minipage}
        \hfill
        \begin{minipage}[b]{0.075\linewidth}
            \centering
            \centerline{\includegraphics[width=\linewidth]{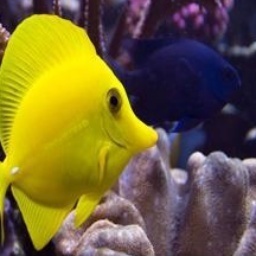}}
        \end{minipage}
    \end{minipage}

    \begin{minipage}[b]{1.0\linewidth}
        \begin{minipage}[b]{0.075\linewidth}
            \centering
            \centerline{\includegraphics[width=\linewidth]{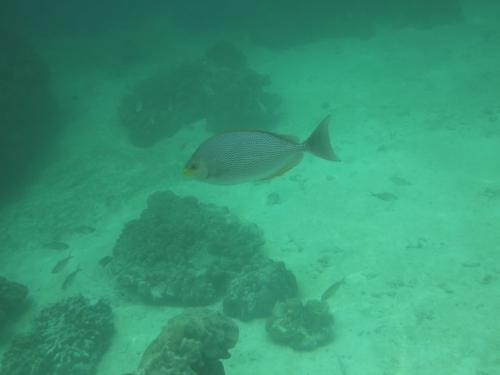}}
            \centerline{(a)}
            \medskip
        \end{minipage}
        \hfill
        \begin{minipage}[b]{0.075\linewidth}
            \centering
            \centerline{\includegraphics[width=\linewidth]{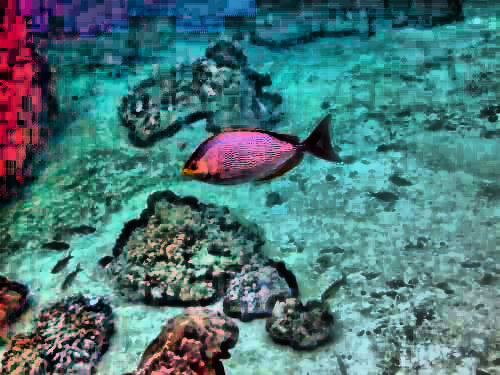}}
            \centerline{(b)}
            \medskip
        \end{minipage}
        \hfill
        \begin{minipage}[b]{0.075\linewidth}
            \centering
            \centerline{\includegraphics[width=\linewidth]{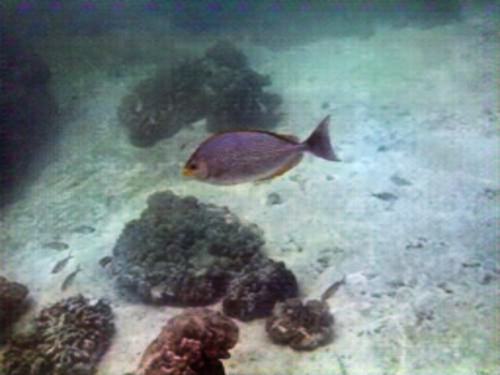}}
            \centerline{(c)}
            \medskip
        \end{minipage}
        \hfill
        \begin{minipage}[b]{0.075\linewidth}
            \centering
            \centerline{\includegraphics[width=\linewidth]{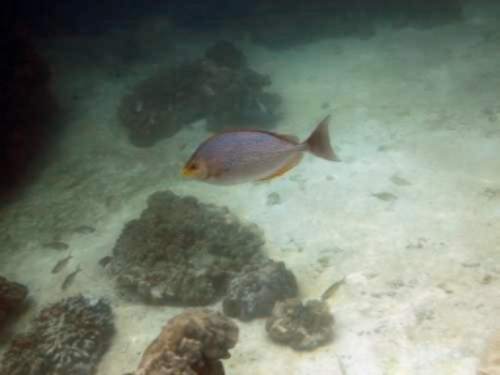}}
            \centerline{(d)}
            \medskip
        \end{minipage}
        \hfill
        \begin{minipage}[b]{0.075\linewidth}
            \centering
            \centerline{\includegraphics[width=\linewidth]{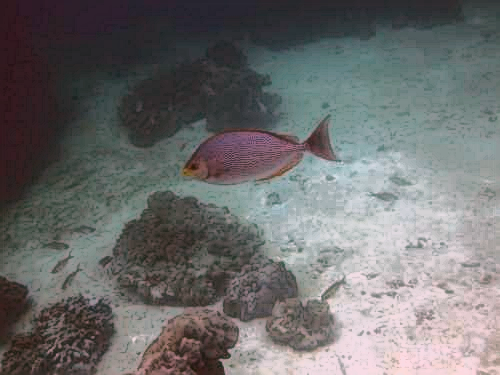}}
            \centerline{(e)}
            \medskip
        \end{minipage}
        \hfill
        \begin{minipage}[b]{0.075\linewidth}
            \centering
            \centerline{\includegraphics[width=\linewidth]{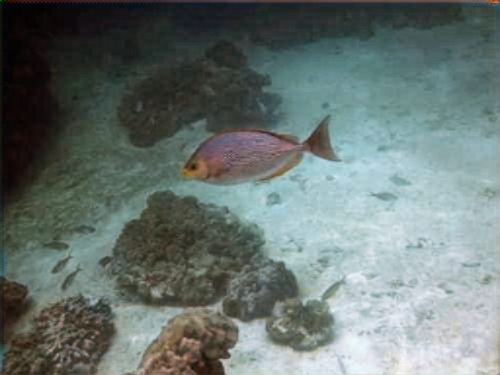}}
            \centerline{(f)}
            \medskip
        \end{minipage}
        \hfill
        \begin{minipage}[b]{0.075\linewidth}
            \centering
            \centerline{\includegraphics[width=\linewidth]{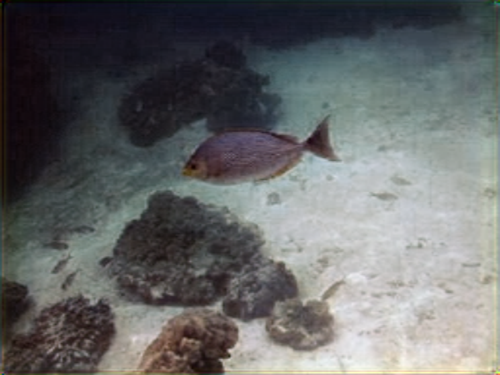}}
            \centerline{(g)}
            \medskip
        \end{minipage}
        \hfill
        \begin{minipage}[b]{0.075\linewidth}
            \centering
            \centerline{\includegraphics[width=\linewidth]{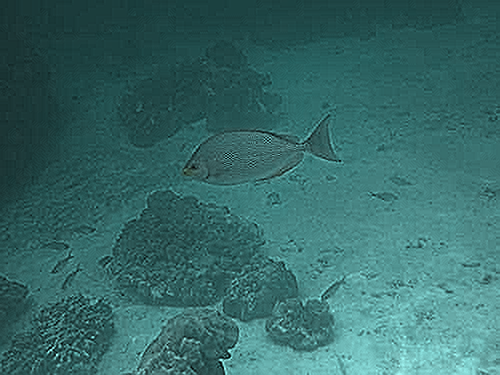}}
            \centerline{(h)}
            \medskip
        \end{minipage}
        \hfill
        \begin{minipage}[b]{0.075\linewidth}
            \centering
            \centerline{\includegraphics[width=\linewidth]{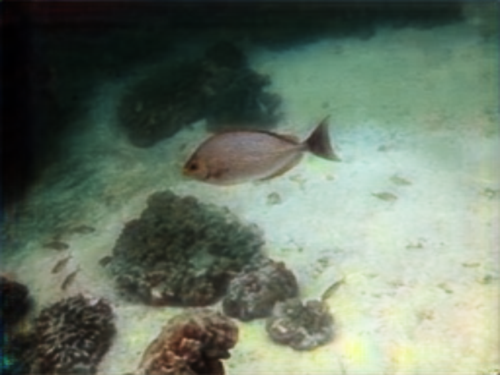}}
            \centerline{(i)}
            \medskip
        \end{minipage}
        \hfill
        \begin{minipage}[b]{0.075\linewidth}
            \centering
            \centerline{\includegraphics[width=\linewidth]{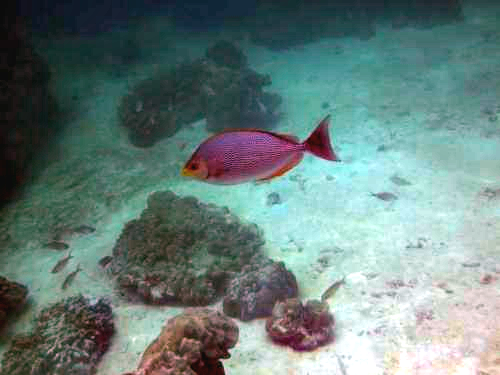}}
            \centerline{(j)}
            \medskip
        \end{minipage}
        \hfill
        \begin{minipage}[b]{0.075\linewidth}
            \centering
            \centerline{\includegraphics[width=\linewidth]{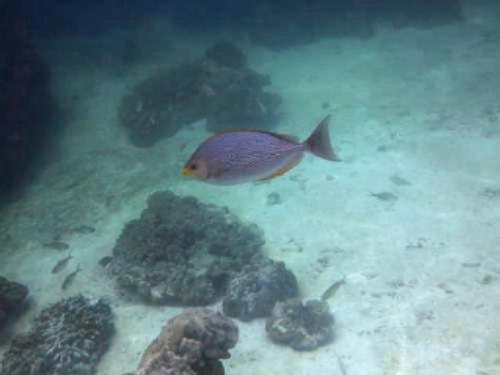}}
            \centerline{(k)}\medskip
        \end{minipage}
        \hfill
        \begin{minipage}[b]{0.075\linewidth}
            \centering
            \centerline{\includegraphics[width=\linewidth]{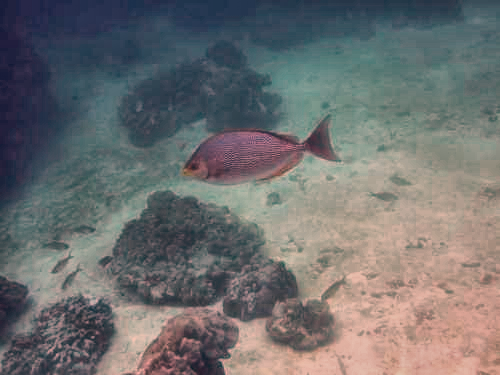}}
            \centerline{(l)}\medskip
        \end{minipage}
    \end{minipage}

    \caption{Qualitative comparison with state-of-the-art methods. From left to right: (a) Input, (b) Sea-thru, (c) FUnIE-GAN, (d) IACC, (e) Bayesian-retinex, (f) TUDA, (g) U-Transformer, (h) ICSP, (i) PUGAN, (j) ROP, (k) MAC-Lookup, and (l) Ground truth.}
    \label{fig:compare}
\end{figure*}
\begin{figure}[ht]
    
    \begin{minipage}[b]{1.0\linewidth}
        \begin{minipage}[b]{0.185\linewidth}
            \centering
            \centerline{\includegraphics[width=\linewidth]{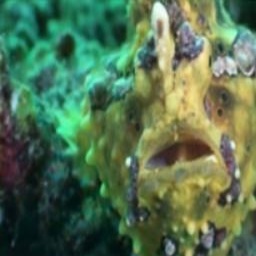}}
        \end{minipage}
        \hfill
        \begin{minipage}[b]{0.185\linewidth}
            \centering
            \centerline{\includegraphics[width=\linewidth]{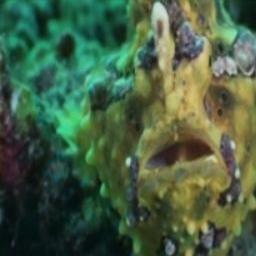}}
        \end{minipage}
        \hfill
        \begin{minipage}[b]{0.185\linewidth}
            \centering
            \centerline{\includegraphics[width=\linewidth]{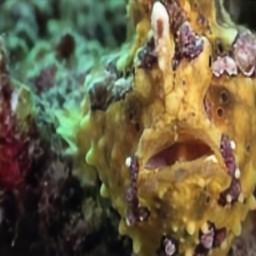}}
        \end{minipage}
        \hfill
        \begin{minipage}[b]{0.185\linewidth}
            \centering
            \centerline{\includegraphics[width=\linewidth]{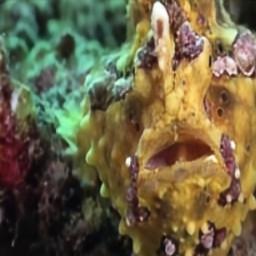}}
        \end{minipage}
        \hfill
        \begin{minipage}[b]{0.185\linewidth}
            \centering
            \centerline{\includegraphics[width=\linewidth]{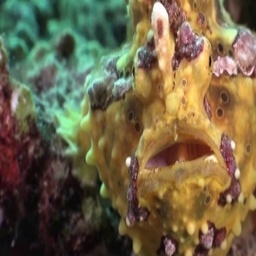}}
        \end{minipage}
    \end{minipage}

    \begin{minipage}[b]{1.0\linewidth}
        \begin{minipage}[b]{0.185\linewidth}
            \centering
            \centerline{\includegraphics[width=\linewidth]{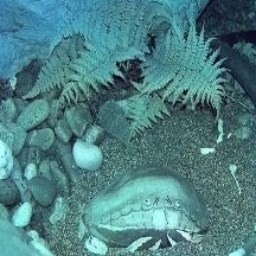}}
        \end{minipage}
        \hfill
        \begin{minipage}[b]{0.185\linewidth}
            \centering
            \centerline{\includegraphics[width=\linewidth]{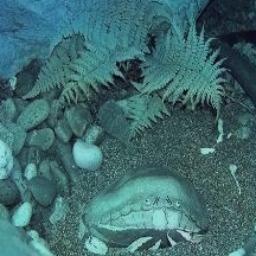}}
        \end{minipage}
        \hfill
        \begin{minipage}[b]{0.185\linewidth}
            \centering
            \centerline{\includegraphics[width=\linewidth]{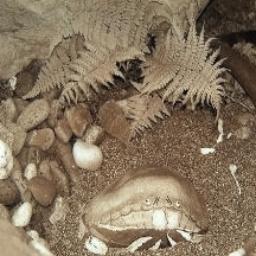}}
        \end{minipage}
        \hfill
        \begin{minipage}[b]{0.185\linewidth}
            \centering
            \centerline{\includegraphics[width=\linewidth]{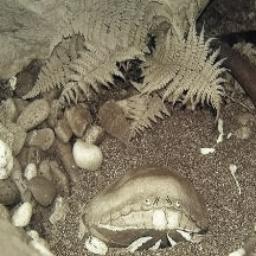}}
        \end{minipage}
        \hfill
        \begin{minipage}[b]{0.185\linewidth}
            \centering
            \centerline{\includegraphics[width=\linewidth]{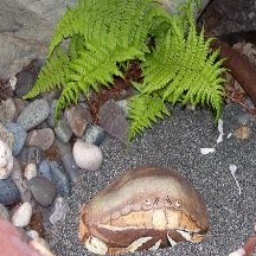}}
        \end{minipage}
    \end{minipage}

    \begin{minipage}[b]{1.0\linewidth}
        \begin{minipage}[b]{0.185\linewidth}
            \centering
            \centerline{\includegraphics[width=\linewidth]{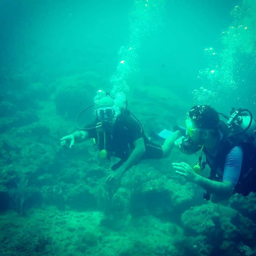}}
            \centerline{(a)}
            \medskip
        \end{minipage}
        \hfill
        \begin{minipage}[b]{0.185\linewidth}
            \centering
            \centerline{\includegraphics[width=\linewidth]{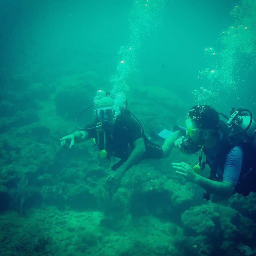}}
            \centerline{(b)}
            \medskip
        \end{minipage}
        \hfill
        \begin{minipage}[b]{0.185\linewidth}
            \centering
            \centerline{\includegraphics[width=\linewidth]{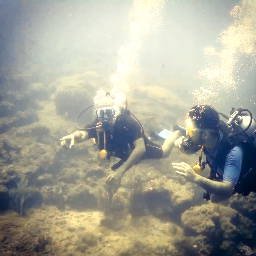}}
            \centerline{(c)}
            \medskip
        \end{minipage}
        \hfill
        \begin{minipage}[b]{0.185\linewidth}
            \centering
            \centerline{\includegraphics[width=\linewidth]{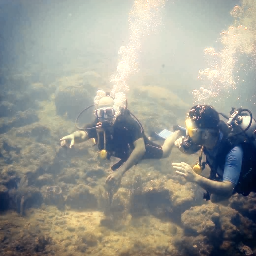}}
            \centerline{(d)}
            \medskip
        \end{minipage}
        \hfill
        \begin{minipage}[b]{0.185\linewidth}
            \centering
            \centerline{\includegraphics[width=\linewidth]{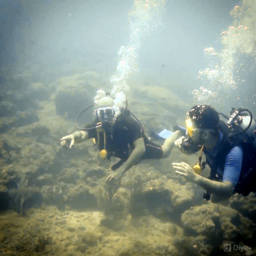}}
            \centerline{(e)}
            \medskip
        \end{minipage}
    \end{minipage}

    \caption{Qualitative ablation studies demonstrating the impact of each key component of our proposed method. The first column presents test images from the UIEB, EUVP, and LSUI datasets. (a) Raw input images. (b) Results without the \textbf{MAAE} module. (c) Results without the \textbf{CLTCC} module. (d) Results using the full MAC-Lookup framework. (e) Ground truth images for reference.}
    \label{fig:ablation}
\end{figure}

\subsection{Visual Comparisons}
As shown in \cref{fig:compare}, traditional methods~\cite{zhuang2021bayesian,wang2023domain,hou2023non,liu2022rank} cause color distortions. Sea-thru and PUGAN inadequately improve visibility, while FUnIE-GAN, IACC, and U-Transformer fail to enhance contrast and face issues like local darkness and over- or under-enhancement.

According to \cref{fig:compare}, Sea-thru, Bayesian-retinex, TUDA, ICSP, and PUGAN struggle with color restoration, making yellow areas appear gray. IACC and U-Transformer restore yellow tones but fail with contrast, while FUnIE-GAN and ROP improve color partially but have contrast issues like local darkness or uneven enhancement.

As shown in \cref{fig:compare}, Sea-thru significantly distorts color in hazy images. ICSP and IACC do not effectively remove blur, reducing visibility enhancement. ROP introduces local darkness, and other methods fail to improve contrast, facing issues like darkness and improper enhancement.

In summary, our method effectively removes unnatural colors, improves visibility, and prevents over- or under-enhancement. According to \cref{tab:experiment}, it achieves the highest PSNR and SSIM, and the lowest LPIPS values, indicating superior contrast, brightness, and texture detail enhancement for natural colors and high visibility.

\subsection{Ablation Studies}
\label{sec:ablation}
We assess the effectiveness of our MAC-Lookup framework through ablation studies on the same datasets, using these setups: (a) without the Multi-Axis Adaptive Enhancement (MAAE) module (-w/o MAAE), (b) without the Conditional 3D Lookup Table Color Correction (CLTCC) module (-w/o CLTCC), and (c) the complete MAC-Lookup framework. \Cref{fig:ablation} shows the qualitative comparisons for these setups on the UIEB, EUVP, and LSUI datasets. The following can be observed:
\begin{itemize}
    \item \textbf{-w/o MAAE}: The absence of the MAAE module leads to suboptimal color correction and contrast enhancement, particularly in scenes with complex lighting conditions, but compare with raw image, the model without MAAE restore the shadow areas and simple color corrections.
    \item \textbf{-w/o CLTCC}: Without the CLTCC module, the method struggles to maintain consistency and color fidelity, resulting in some image distortion. However, compared to the original image, it has effectively restored the overall color and structure, with natural color transitions and improved contrast in shadowed areas.
    \item \textbf{MAC-Lookup (full)}: The full model, incorporating both MAAE and CLTCC modules, achieves the most visually pleasing results, with  color restoration, enhanced contrast, and preserved texture details.
\end{itemize}

\begin{table}[ht]
\centering
\caption{Quantitative average results of ablation studies on the three datasets. \checkmark\ and \ding{55} indicate the presence and absence of each component, respectively.}
\label{tab:ablation}
\begin{tabular}{ccc|ccc}
\toprule
\multirow{2}{*}{Index} & \multirow{2}{*}{CLT} & \multirow{2}{*}{MAGM} & \multicolumn{3}{c}{Metrics} \\ 
\cmidrule(lr){4-6} 
& & & PSNR$\uparrow$ & SSIM$\uparrow$ & LPIPS$\downarrow$ \\ 
\midrule
(1) & \ding{55} & \checkmark & 25.74 & 0.9123 & 0.1229 \\
(2) & \checkmark & \ding{55} & 18.20 & 0.7793 & 0.3076 \\
(3) & \checkmark & \checkmark & 26.79 & 0.9147 & 0.1159 \\ 
\bottomrule
\end{tabular}
\end{table}

In \cref{tab:ablation}, we evaluate ablated models across three datasets, showing that the complete MAC-Lookup framework consistently surpasses the ablated versions in PSNR, SSIM, and LPIPS metrics. This highlights each component's importance to our method's overall performance. The CLTCC module is crucial for preserving shadows and colors, while the MAAE module improves adaptability to different underwater conditions and maintains color consistency.

\section{Conclusion}
The MAC-Lookup model introduced here enhances underwater imagery by combining the Conditional 3D Lookup Table (CLTCC) with Multi-Axis Adaptive Enhancement (MAAE). It addresses common issues like color distortion, blur, and uneven contrast. The conditional vector, derived from comparing ground truth with CLTCC-processed images, forms the basis for initial corrections, while MAAE refines quality by tackling over-enhancement and over-saturation, improving image details. Experiments show its performance aligns with the state-of-the-art. Future work will focus on adapting to changing conditions and adding data sources to improve training. This approach aims for deeper insights into underwater environments to aid conservation.
\bibliographystyle{IEEEbib}
\bibliography{ref}

\end{document}